%% file: main.tex
\documentclass[11pt]{article}

\usepackage[margin=1in]{geometry}
\usepackage{times}

\input{math_commands.tex}

\usepackage{hyperref}
\hypersetup{
    hidelinks
}
\usepackage{url}

\title{RefineDrive: Reliable Failure-Guided Learning for Vision-Language-Action Driving}

\usepackage{graphicx}
\usepackage{booktabs}
\usepackage[export]{adjustbox}
\usepackage{amsmath}
\usepackage{amssymb}

\author{
    Zhe Sun$^{1}$, Ziyi Luo$^{1}$, Yehao Lu$^{1}$,
    Lei Zhou$^{2}$, and Xi Li$^{1}$\thanks{Corresponding author: \texttt{xilizju@zju.edu.cn}}
    \\[0.5em]
    \small $^{1}$College of Computer Science and Technology,
    Zhejiang University, Hangzhou, China
    \\
    \small $^{2}$Yinwang Intelligent Technology Co., Ltd.
}

\date{}

\begin{document}

\maketitle

\begin{abstract}
Vision-Language-Action (VLA) models for autonomous driving rely heavily on successful expert demonstrations, leaving model-specific failures underexploited.
Learning from these failures is hindered by unreliable diagnoses, poorly matched correction targets, and coarse rewards.
We propose \textbf{RefineDrive}, a failure-guided post-training framework that learns from self-generated failures through targeted supervision and safety-aware reinforcement learning.
Reliable Diagnosis derives structured, verifiable feedback on collisions and drivable-area violations directly from simulator states.
Minimum-Correction Target Retrieval searches a clustered human trajectory bank for nearby corrections that satisfy hard-safety constraints in the current scene, prioritizing preservation of the failed prediction's motion pattern.
Conditioned on the driving context and failed trajectory, Correction SFT learns to generate the diagnosis followed by the retrieved correction as a training-only auxiliary task.
We then apply GRPO with a Safety-Layered Reward that strictly prioritizes hard-safe trajectories, retains continuous safety feedback for both unsafe and hard-safe trajectories, and rewards driving progress only after hard safety is satisfied.
At inference, the policy directly predicts trajectories from the driving context without an explicit diagnosis or repair stage.
On NAVSIM v1, RefineDrive improves the 4B base SFT policy from 87.7 to 91.7 PDMS.
Using the same checkpoint without additional training, RefineDrive achieves 89.4 EPDMS on the original NAVTEST scenes evaluated with NAVSIM v2 extended metrics.
Controlled ablations support the benefits of structured diagnosis supervision, retrieved corrections, and safety-layered optimization for direct planning.
\end{abstract}

\begin{figure}[t]
    \centering
    \setlength{\belowcaptionskip}{3pt}
    \includegraphics[width=\linewidth]{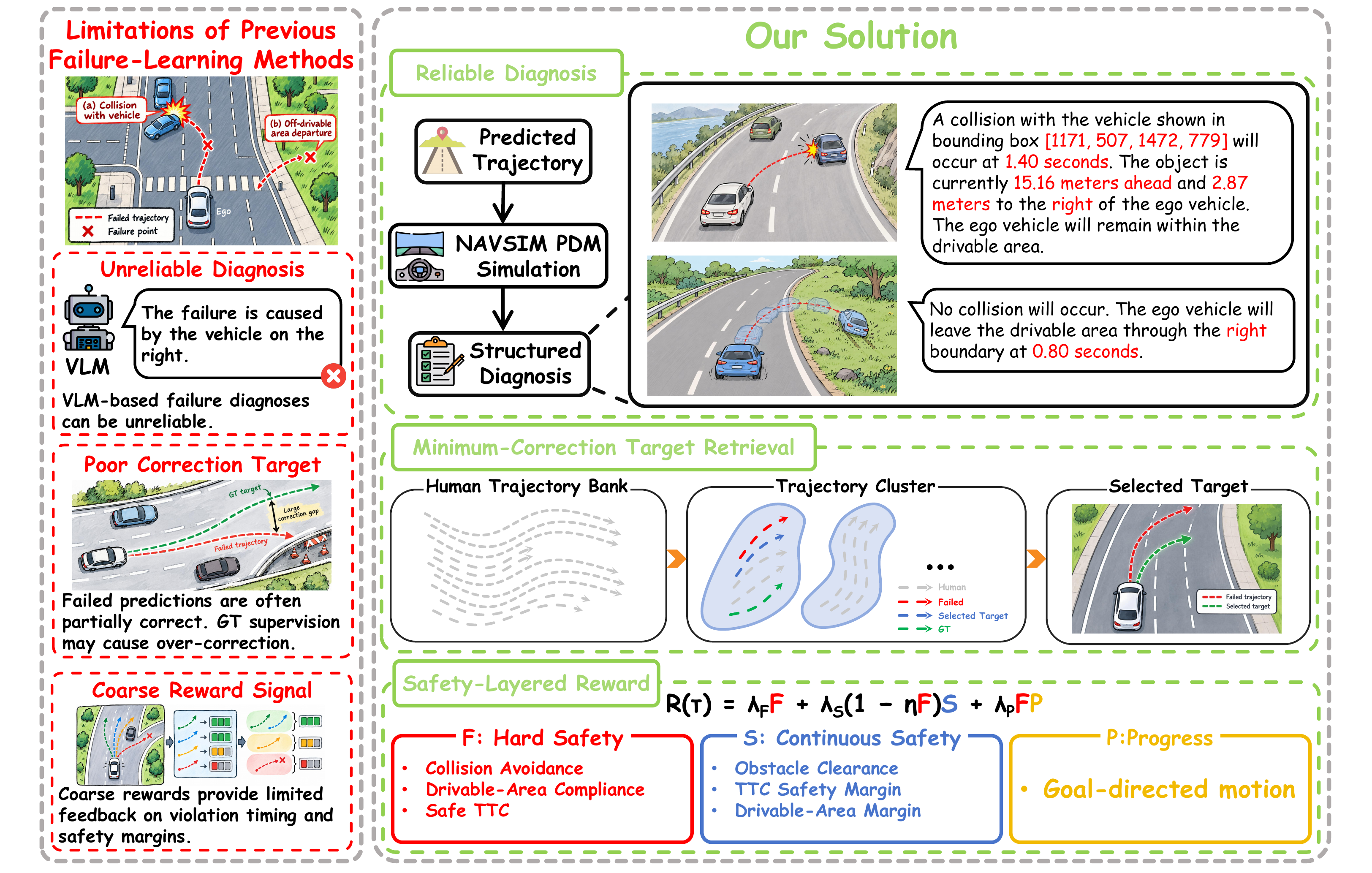}
    \caption{\textbf{Motivation and overview of RefineDrive.} Existing failure-learning methods can suffer from unreliable VLM-based diagnoses, over-corrective human ground-truth supervision, and coarse aggregated rewards. We address these limitations with Reliable Diagnosis, which grounds structured feedback in NAVSIM PDM simulation; Minimum-Correction Target Retrieval, which searches a clustered human trajectory bank for safe corrections close to failed predictions; and Safety-Layered Reward, which provides hierarchical signals over hard safety, continuous safety, and driving progress.}
    \label{fig:intro}
\end{figure}
\input{sections/introduction}
\input{sections/related_work}
\begin{figure*}[h]
    \centering
    \setlength{\belowcaptionskip}{3pt}
    \includegraphics[width=\textwidth]{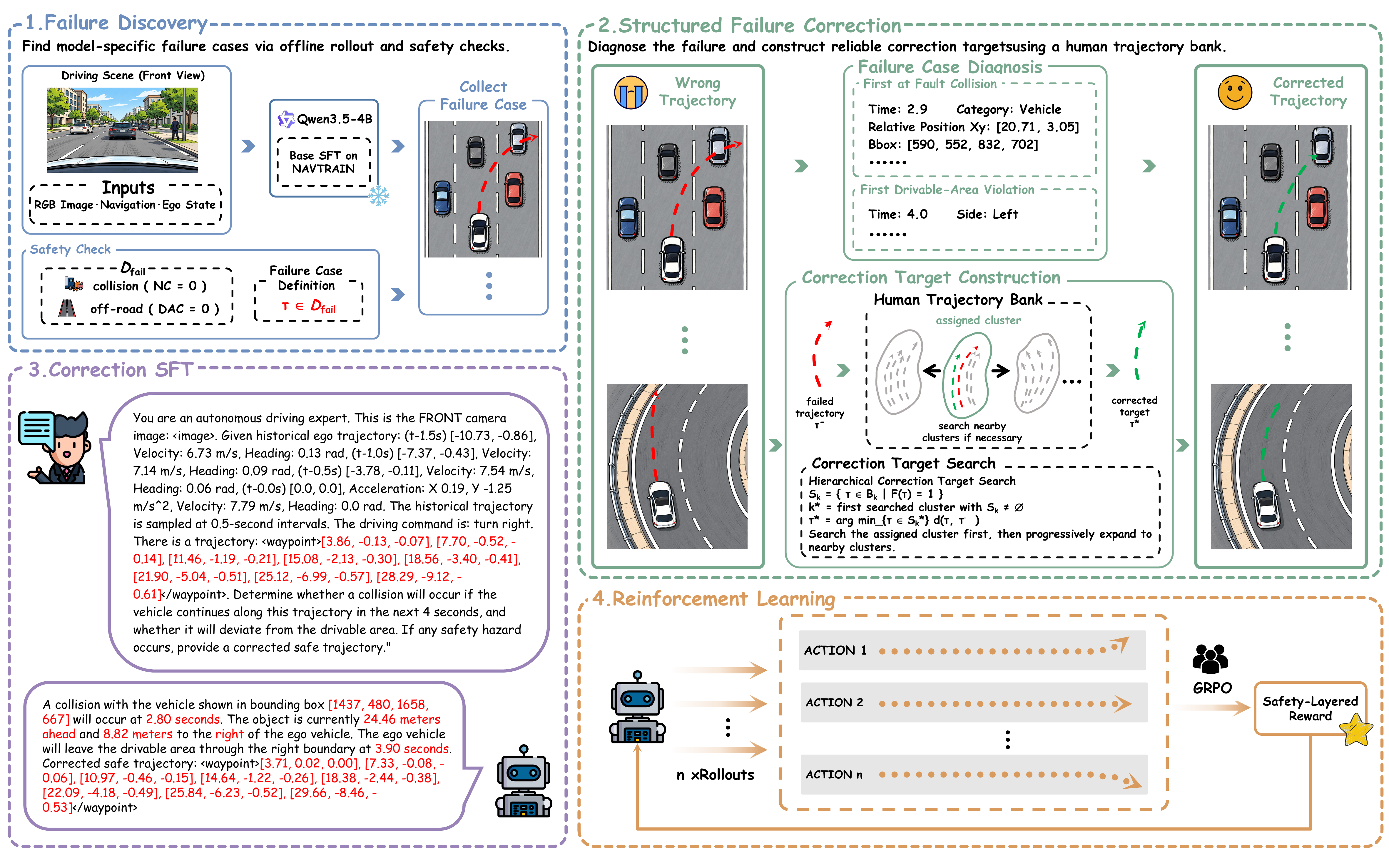}
    \caption{
        \textbf{Training pipeline of RefineDrive.}
        Offline rollouts and simulator-based safety checks expose model-specific
        failures. For each failure, we construct a structured diagnosis and
        retrieve a nearby safe correction from a human trajectory bank.
        Conditioned on the driving context and failed trajectory, Correction SFT
        learns to generate the diagnosis followed by the correction.
        Finally, GRPO refines the policy with the Safety-Layered Reward,
        providing fine-grained feedback beyond aggregated driving scores.
        Diagnosis and correction are training-only tasks; inference directly maps the driving context to the final trajectory.
    }
    \label{fig:method}
\end{figure*}
\input{sections/method}
\input{sections/experiment}
\input{sections/conclusion}

\newpage
\bibliographystyle{unsrt}
\bibliography{references}

\end{document}

%% file: math_commands.tex
\usepackage{amsmath,amsfonts,bm}

\def\eqref#1{equation~\ref{#1}}

\def\1{\bm{1}}

\DeclareMathAlphabet{\mathsfit}{\encodingdefault}{\sfdefault}{m}{sl}
\SetMathAlphabet{\mathsfit}{bold}{\encodingdefault}{\sfdefault}{bx}{n}



%% file: sections/introduction.tex
\section{Introduction}
\label{sec:introduction}

Vision-Language-Action (VLA) models have shown strong potential for end-to-end autonomous driving, benefiting from large-scale vision-language pretraining and expert demonstrations. However, most approaches still rely primarily on imitation learning from successful human behavior, while failures exposed by the model's own rollouts---such as collisions and departures from the drivable area---remain underexploited. For a policy with strong basic driving capabilities, further improvement may depend less on imitating more expert behavior and more on identifying and correcting the remaining weaknesses revealed by its own failures.

Recent studies have begun to explore learning from failures, but three key limitations remain, as illustrated in Fig.~\ref{fig:intro}.
First, \textbf{failure diagnosis can be unreliable}. Methods that rely on large VLMs or teacher models to explain failures may produce open-ended diagnoses that are difficult to verify, with inaccurate event localization, incorrect causal attribution, or inconsistency with the simulated outcome.
Second, \textbf{the correction target may be poorly matched to the failure}. Since a failed trajectory is often only partially incorrect, directly replacing it with human ground truth can introduce unnecessary behavioral changes and lead to over-correction.
Third, \textbf{coarse reward signals provide limited optimization guidance}. An aggregated driving score may insufficiently distinguish the extent and timing of safety violations among unsafe trajectories or safety margins among safe ones, limiting fine-grained safety optimization.

To address these limitations, we propose \textbf{RefineDrive}, a failure-guided
post-training approach with three components.
\textbf{Reliable Diagnosis} derives structured, simulator-verifiable
feedback for collisions and drivable-area violations.
\textbf{Minimum-Correction Target Retrieval} searches a clustered human
trajectory bank for a nearby correction that satisfies hard-safety
constraints in the current scene, prioritizing the motion pattern of
the failed prediction.
During training, Correction SFT conditions on the driving context
and a failed trajectory to learn structured diagnosis and correction.
At inference, RefineDrive directly predicts the final trajectory
from the driving context, without an explicit diagnosis or repair stage.

Finally, \textbf{Safety-Layered GRPO} combines hard safety, continuous
safety, and driving progress.
It strictly prioritizes hard-safe trajectories, retains continuous
safety feedback for both unsafe and hard-safe trajectories, and rewards
progress only after hard safety is satisfied.

RefineDrive uses self-generated rollouts to expose residual weaknesses,
derives simulator-verifiable diagnoses, retrieves minimally modified safe
corrective targets, and further refines the policy with safety-layered
reinforcement learning. Rather than repeatedly learning the full
distribution of expert behavior, RefineDrive focuses post-training on
the specific deficiencies currently exhibited by the policy.

Our main contributions are summarized as follows:
\begin{itemize}
    \item We introduce \textbf{RefineDrive}, which explicitly exploits
    failures from the model's own rollouts to address residual weaknesses
    in autonomous driving VLAs.

    \item We propose \textbf{Reliable Diagnosis and Minimum-Correction
    Target Retrieval}. Simulator-derived structured diagnoses provide
    reproducible and verifiable failure feedback, while cluster-based
    retrieval selects safe corrective trajectories close to failed
    predictions, reducing unnecessary behavioral changes and
    over-correction.

    \item We develop a \textbf{Safety-Layered Reward} that prioritizes
    hard safety, provides continuous safety feedback for both unsafe
    and hard-safe trajectories, and rewards driving progress only after
    hard safety is satisfied. Experiments on NAVSIM demonstrate the
    effectiveness of RefineDrive.
\end{itemize}

%% file: sections/related_work.tex
\section{Related Work}

\subsection{Vision-Language-Action Models for Autonomous Driving}

Vision-language models (VLMs) have been applied to autonomous driving for scene understanding, reasoning, and decision making~\cite{xu2024drivegpt4,mao2023gpt,shao2024lmdrive,sima2024drivelm,nie2024reason2drive}. DriveVLM~\cite{tian2024drivevlm} emphasizes language-space reasoning for hierarchical planning, whereas VLA models more tightly integrate multimodal reasoning and action generation. OpenDriveVLA~\cite{zhou2026opendrivevla} aligns instance-aware visual and language representations for autoregressive actions, while AutoVLA~\cite{zhou2026autovla} combines semantic reasoning, discretized action tokens, and reinforcement fine-tuning. ReCogDrive~\cite{xiong2026recogdrive} integrates a cognitive VLM with a diffusion planner, and Qwen-Drive~\cite{zhou2026qwen} explores unified representations for perception, understanding, and planning.

ReflectDrive~\cite{li2026discrete} introduces discrete-diffusion planning with inference-time safety reflection: a safety scorer identifies unsafe waypoints, local search finds safe anchors, and inpainting regenerates the surrounding trajectory. Although both methods repair unsafe predictions, ReflectDrive performs waypoint-level inference editing, whereas we retrieve complete, scene-validated trajectories and pair them with structured diagnoses as offline Correction SFT supervision.

\subsection{Learning from Failures}

Learning from failures complements imitation from successful demonstrations. ChauffeurNet~\cite{bansal2018chauffeurnet} perturbs expert trajectories to synthesize unsafe situations. Recent VLA methods exploit failures more directly: ELF-VLA~\cite{luo2026unleashing} uses teacher-generated diagnoses for trajectory refinement, SafeAlign-VLA~\cite{tian2026safealign} constructs failure descriptions and counterfactual positives for supervised and reinforcement learning, and FIRE-VLA~\cite{dou2026fire} employs failure-triggered self-distillation with privileged future information. Rather than relying on open-ended teacher explanations, we derive structured diagnoses directly from simulator states, grounding violation times, involved objects, and boundary-exit directions in verifiable geometric and kinematic evidence.

The concurrent R$^2$LPL framework~\cite{gong2026learning} retrieves feasible trajectory anchors from recoverable closed-loop states and
updates an anchor-scoring planner through replay-based lifelong learning. Its target scores combine rule-based planning quality with route and expert consistency using state-dependent weights. Both methods construct corrective supervision from policy failures, but differ in the correction setting and target-selection criterion. We formulate our correction stage as trajectory-conditioned behavior repair as a training-only auxiliary task: the failed prediction serves as both the reference for retrieving a nearby, scene-verified safe trajectory from a clustered human trajectory bank and an explicit conditioning input for Correction SFT. Rather than supervising anchor scores, we train the VLA to autoregressively generate a simulator-grounded diagnosis followed by the corrected trajectory. Safety-Layered GRPO subsequently refines the policy.

%% file: sections/method.tex
\section{Method}
\label{sec:method}

\subsection{Overview}
\label{sec:overview}

We propose RefineDrive that enables an already capable driving VLA to improve from safety-critical failures exposed by its own behavior.
As illustrated in Fig.~\ref{fig:method}, our framework consists of four stages:
\textbf{(1) Failure Discovery},
\textbf{(2) Structured Failure Correction},
\textbf{(3) Correction SFT},
and \textbf{(4) Safety-Layered Reinforcement Learning}.

The resulting framework progressively improves the policy by focusing learning on weaknesses revealed by its own rollouts rather than repeatedly imitating already-mastered expert behavior.

\subsection{Failure Discovery}
\label{sec:failure_discovery}

We start from a supervised fine-tuned driving policy and perform offline rollouts on the training scenarios.
Each predicted trajectory is re-executed using the NAVSIM PDM Simulator to obtain reproducible safety outcomes.

We focus on two directly observable safety-critical failures:
\textbf{at-fault collisions} and \textbf{departures from the drivable area}.
Accordingly, we construct the model-specific failure set as
\begin{equation}
\mathcal{D}_{\mathrm{fail}}
=
\left\{
(q,\tau^{-})
\;\middle|\;
\mathrm{NC}(\tau^{-})=0
\;\lor\;
\mathrm{DAC}(\tau^{-})=0
\right\},
\label{eq:failure_discovery}
\end{equation}
where $q$ denotes the driving context, $\tau^{-}$ is the trajectory predicted by the current policy, and NC and DAC denote No At-Fault Collisions and Drivable Area Compliance, respectively.

Unlike artificially synthesized negative samples, the trajectories in $\mathcal{D}_{\mathrm{fail}}$ expose failure modes actually produced by the current policy.
They therefore provide targeted cases for subsequent diagnosis and correction.

\subsection{Structured Failure Correction}
\label{sec:structured_failure_correction}

For each discovered failure, we construct correction supervision that answers two complementary questions:
\emph{why does the current prediction fail?}
and
\emph{what is the smallest behavioral change that makes it safe?}
We address them through verifiable failure diagnosis and counterfactual minimal correction.

\paragraph{Verifiable Failure Diagnosis.}

Rather than relying on a teacher VLM to produce open-ended explanations, we derive failure feedback directly from simulator states.

For an at-fault collision, we identify the \textbf{first ego at-fault collision} and extract its occurrence time, object category, and relative position with respect to the ego vehicle.
The corresponding traffic participant is further projected onto the current front-view image to obtain its 2D bounding box.
The resulting feedback explicitly indicates when the collision occurs, which object is involved, and where the object appears in the current visual observation.

For a drivable-area violation, we identify the first time at which the predicted trajectory leaves the valid driving region and determine whether the ego vehicle exits through the left or right boundary.

When the collision object cannot be reliably localized in the current visual observation, we discard the corresponding sample to avoid introducing textual supervision that cannot be grounded in the model input.

The resulting diagnosis is therefore derived from reproducible geometric and kinematic verification rather than subjective model interpretation.

\paragraph{Counterfactual Minimal Correction.}

For an already capable driving policy, a failed trajectory's maneuver intention and most of its future motion may remain reasonable, with only a limited deviation causing the safety violation.
Direct replacement with the human trajectory may introduce unnecessary behavioral changes; we instead search for a \textbf{nearby feasible correction} from a real human trajectory bank.

We construct the bank from the training set and group trajectories by motion pattern using K-Means.
Given a failed trajectory $\tau^{-}$, we identify its assigned cluster in the same feature space and search it first to preserve the original motion pattern.
If no feasible correction is found, we search the remaining clusters in ascending order of the distance from their centers to the feature representation of $\tau^{-}$.

Importantly, human trajectories are not regarded as valid correction targets by default.
Since trajectories in the bank may originate from different scenes, every candidate is re-evaluated in the current failure scene.
We define hard-safety feasibility as
\begin{equation}
    F(\tau) =
    \mathbb{I}\!\left[
        \mathrm{NC}(\tau)=1
        \land \mathrm{DAC}(\tau)=1
        \land \mathrm{TTC}(\tau)=1
    \right],
    \label{eq:hard_safety}
\end{equation}
where NC, DAC, and TTC denote no-at-fault collision, drivable-area compliance, and time-to-collision safety, respectively.

To measure deviation from the original prediction, we use the same trajectory representation as K-Means clustering.
We unwrap each trajectory's heading sequence and express it relative to the first waypoint's heading.
Longitudinal positions, lateral positions, and relative headings are normalized using trajectory-bank statistics and flattened across all future waypoints into $\phi(\tau)$.
The trajectory distance is
\begin{equation}
    d(\tau_i,\tau_j)
    = \left\|\phi(\tau_i)-\phi(\tau_j)\right\|_2.
    \label{eq:trajectory_distance}
\end{equation}

Let $\mathcal{B}_k$ denote the set of human trajectories belonging to cluster $k$.
For each searched cluster, we define its feasible candidate set as
\begin{equation}
    \mathcal{S}_k
    = \{\tau \in \mathcal{B}_k \mid F(\tau)=1\}.
    \label{eq:feasible_set}
\end{equation}
If $\mathcal{S}_k$ is non-empty, the closest feasible trajectory within that cluster is selected as
\begin{equation}
    \tau_k^{*}
    = \arg\min_{\tau \in \mathcal{S}_k}
      d(\tau,\tau^{-}).
    \label{eq:cluster_correction}
\end{equation}
The final correction target $\tau^{*}$ is given by $\tau_k^{*}$ from the first searched cluster that contains at least one feasible candidate.
In practice, candidates within each cluster are evaluated in ascending order of $d(\tau,\tau^{-})$, so the search terminates as soon as the first safe candidate is found.

If no feasible trajectory is found after all available candidates have been considered, or if the predefined search budget is reached before finding one, no corrected trajectory is assigned to the corresponding failure case.

\subsection{Correction SFT}
\label{sec:correction_sft}
We use Correction SFT as a training-only auxiliary task that exposes
the policy to self-generated failures, their simulator-grounded diagnoses,
and corresponding safe corrections.

For each training sample, the input contains the driving context $q$
and the failed trajectory $\tau^{-}$.
The target response $y^{*}$ consists of the structured diagnosis followed
by the retrieved corrective trajectory $\tau^{*}$.
We optimize the policy using the standard autoregressive objective:
\begin{equation}
\mathcal{L}_{\mathrm{corr}}
=
-
\sum_{t=1}^{|y^{*}|}
\log
\pi_{\theta}
\left(
y^{*}_{t}
\mid
q,\tau^{-},y^{*}_{<t}
\right).
\label{eq:correction_sft}
\end{equation}
The purpose of this task is to improve direct planning through
failure-specific supervision, rather than to define an inference-time
repair procedure.
At inference, the policy directly generates the final trajectory
conditioned only on the driving context $q$.
It is not given a failed trajectory and does not perform an explicit
diagnosis-and-correction pass.

After Correction SFT, we further refine the policy with
Safety-Layered GRPO.

\subsection{Safety-Layered Reward}
\label{sec:safety_grpo}

We refine the corrected policy with GRPO using a reward that combines
\emph{hard safety}, \emph{continuous safety assessment}, and
\emph{driving progress}.
Hard-safety feasibility is determined by $F(\tau)\in\{0,1\}$ in Eq.~\ref{eq:hard_safety}, which requires NC, DAC, and TTC to be satisfied simultaneously.

\paragraph{Continuous Safety Score.}
Binary feasibility alone cannot distinguish failure severity among
unsafe trajectories or safety margins among feasible ones.
We therefore construct a continuous safety score from simulator-derived
margins and violation times.
For each safety factor
$k\in\{\mathrm{clr},\mathrm{ttc},\mathrm{road}\}$,
corresponding to obstacle clearance, TTC safety, and drivable-area
compliance, let $m_k(\tau)$ denote its minimum signed margin and
$t_k(\tau)$ its first violation time.
We normalize them as
\begin{equation}
\bar m_k(\tau)
=
\operatorname{clip}
\left(
\frac{m_k(\tau)-l_k}{u_k-l_k},0,1
\right),
\qquad
\bar t_k(\tau)
=
\operatorname{clip}
\left(
\frac{t_k(\tau)}{T},0,1
\right),
\label{eq:safety_normalization}
\end{equation}
where $[l_k,u_k]$ is the factor-specific normalization interval
and $T$ is the simulation horizon.
We set $t_k(\tau)=T$ when the corresponding violation does not occur.
The continuous safety score is
\begin{equation}
S(\tau)
=
\sum_k w_k
\left[
(1-\rho_k)\bar m_k(\tau)
+
\rho_k\bar t_k(\tau)
\right],
\qquad
\sum_k w_k=1,
\label{eq:continuous_safety}
\end{equation}
where $w_k\geq0$ and $\rho_k\in[0,1]$, ensuring $S(\tau)\in[0,1]$.
Higher scores favor larger safety margins and later violations.
Thus, $S$ provides graded feedback on failure severity for unsafe
trajectories while continuing to distinguish safety margins among
hard-safe trajectories.

\paragraph{Safety-Layered Composition.}
We reward driving progress only after hard safety is satisfied.
The progress score is defined using Ego Progress (EP):
\begin{equation}
P(\tau)
=
\mathrm{EP}(\tau)
\label{eq:progress_score}
\end{equation}
The final reward is
\begin{equation}
R(\tau)
=
\lambda_F F(\tau)
+
\lambda_S\bigl[1-\eta F(\tau)\bigr]S(\tau)
+
\lambda_P F(\tau)P(\tau),
\label{eq:safety_layered_reward}
\end{equation}
where $\eta\in[0,1]$ controls the attenuation of the continuous
safety reward once hard safety is satisfied.
Unsafe trajectories receive only $\lambda_S S(\tau)$.
Hard-safe trajectories receive the feasibility bonus, retain a
reduced safety-score weight of $\lambda_S(1-\eta)$, and additionally
receive the progress reward.

We set $(\lambda_F,\lambda_S,\lambda_P)=(2.0,1.0,0.25)$ and
$\eta=0.5$, retaining half of the continuous safety weight for
hard-safe trajectories.
Consequently, unsafe trajectories have $R(\tau)\in[0,1]$, whereas
hard-safe trajectories have $R(\tau)\in[2,2.75]$.
This guarantees a strict reward preference for hard-safe trajectories
while preserving fine-grained safety feedback within both groups.

%% file: sections/experiment.tex
\section{Experiments}
\label{sec:experiments}

\subsection{Experimental Setup}
\label{sec:exp_setup}

\paragraph{Dataset and Evaluation Protocol.}
We conduct experiments on NAVSIM~\cite{navsim}, a planning-oriented autonomous driving benchmark built on OpenScene/nuPlan.
Following the official NAVSIM v1 protocol, we use NAVTRAIN for supervised training, failure mining, and reinforcement learning, and reserve NAVTEST exclusively for evaluation.
All corrective trajectories are constructed from the training split only.
We additionally evaluate the same checkpoint used for NAVSIM v1
on the original NAVTEST scenes using the extended metrics of NAVSIM v2~\cite{Cao2025CORL},
without additional training.
This is a single-stage, original-scene evaluation and does not include
synthetic second-stage observations or two-stage pseudo-simulation
aggregation.

\paragraph{Metrics.}
We report Predictive Driver Model Score (PDMS) for overall
benchmark comparison. To evaluate our safety-oriented
objective, we analyze No At-Fault Collision (NC),
Drivable Area Compliance (DAC), and Time-to-Collision (TTC)
as the key safety outcomes, alongside Ego Progress (EP)
and Comfort (C). We interpret safety and progress jointly:
similar aggregate PDMS values need not imply similar
safety performance. All metrics are higher-is-better.

\paragraph{Model and Training Details.}
We instantiate RefineDrive with Qwen3.5-4B and first perform trajectory
SFT on NAVTRAIN.
The resulting policy is rolled out to mine safety-critical failures,
from which we construct the diagnosis-and-correction dataset.
We then perform Correction SFT followed by Safety-Layered GRPO.
All ablations start from the same base SFT checkpoint.
During evaluation, every checkpoint, including those evaluated
immediately after Correction SFT, directly predicts eight
$(x,y,\mathrm{heading})$ waypoints at $0.5$\,s intervals over a
$4$\,s horizon from the current front-view image, ego state, and
historical trajectory.
No failed trajectory is supplied, and no additional diagnosis or trajectory-repair pass is performed.

\subsection{Main Results}
\label{sec:main_results}

\paragraph{Results on NAVSIM v1.}
Table~\ref{tab:navsim_v1_main} compares our method with representative end-to-end and VLA planners on NAVSIM v1.
Our 4B model achieves $91.7$ PDMS, together with $98.7$ NC, $98.2$ DAC, and $95.9$ TTC, demonstrating strong overall and safety performance without relying on a substantially larger foundation model.

\begin{table}[t]
    \centering
    \setlength{\belowcaptionskip}{6pt}
    \caption{
        Comparison with state-of-the-art methods on NAVSIM v1.
        All metrics are higher-is-better.
    }
    \label{tab:navsim_v1_main}
    \small
    \resizebox{\linewidth}{!}{%
    \begin{tabular}{lccccccc}
        \toprule
        Method & Params. & NC & DAC & TTC & C & EP & PDMS \\
        \midrule
        \multicolumn{8}{l}{\textit{End-to-End Methods}} \\
        \textit{UniAD~\cite{uniad}}
            & -- & 97.8 & 91.9 & 92.9 & 100 & 78.8 & 83.4 \\
        \textit{TransFuser~\cite{transfuser}}
            & -- & 97.7 & 92.8 & 92.8 & 100 & 79.2 & 84.0 \\
        \textit{DiffusionDrive~\cite{diffusiondrive}}
            & -- & 98.2 & 96.2 & 94.7 & 100 & 82.2 & 88.1 \\
        \midrule
        \multicolumn{8}{l}{\textit{Driving VLA Methods}} \\
        \textit{Base SFT (Qwen3.5-4B)~\cite{qwen3.5}}
            & 4B & 98.6 & 95.8 & 95.4 & 100 & 81.1 & 87.7 \\
        \textit{AutoVLA~\cite{zhou2026autovla}}
            & 3B & 98.4 & 95.6 & 98.0 & 99.9 & 81.9 & 89.1 \\
        \textit{SafeAlign-VLA~\cite{tian2026safealign}} & 7B & 98.6 & 97.2 & 98.1 & 100 & 81.7 & 89.1 \\
        \textit{DriveVLA-W0~\cite{drivevla}}
            & 8B & 98.7 & 99.1 & 95.3 & 99.3 & 83.3 & 90.2 \\
        \textit{ReCogDrive~\cite{xiong2026recogdrive}}
            & 2B & 97.9 & 97.3 & 94.9 & 100 & 87.3 & 90.8 \\
        \textit{ELF-VLA~\cite{luo2026unleashing}}
            & 8B & 98.9 & 98.1 & 96.0 & 100 & 85.3 & 91.0 \\
        \textit{DriveTeach-VLA~\cite{drive-teach-vla}}
            & 3B & 98.5 & 96.9 & 97.9 & 98.2 & 88.5 & 90.4 \\
        \textit{DriveMA~\cite{drivema}}
            & 4B & 98.7 & 97.8 & 95.5 & 99.9 & 86.7 & 91.2 \\
        \textit{Qwen-Drive-1.0-RL~\cite{zhou2026qwen}}
            & 4B & 98.6 & 98.2 & 95.9 & 100 & 84.8 & 90.7 \\
        \textit{ReflectDrive~\cite{li2026discrete}}
            & 8B & 97.7 & 99.3 & 93.5 & 100 & 86.9 & 91.1 \\
        \textit{DriveFine~\cite{dang2026drivefine}}
            & 8B & 98.6 & 97.9 & 95.2 & 99.9 & 85.5 & 90.7 \\
        \midrule
        \textbf{RefineDrive}
            & 4B & 98.7 & 98.2 & 95.9 & 100 & 87.2 & 91.7 \\
        \bottomrule
    \end{tabular}%
    }
\end{table}

\paragraph{Extended-Metric Evaluation on NAVTEST.}
Using the extended metrics of NAVSIM v2 on the original NAVTEST scenes,
RefineDrive achieves 89.4 EPDMS with the same checkpoint used for
NAVSIM v1 and no additional training.
This complements the v1 results with a broader assessment of planning
quality on the same test split.

\begin{table}[t]
    \centering
    \setlength{\belowcaptionskip}{6pt}
    \caption{
        Comparison on the original NAVTEST scenes using NAVSIM v2
        extended metrics.
        Our result uses the same checkpoint as the NAVSIM v1 evaluation,
        without additional training.
        All metrics are higher-is-better.
        * indicates the exact NAVSIM v2 evaluator revision is not specified in the paper.
        † indicates results evaluated on the bug-fixed version of NAVSIM.
    }
    \label{tab:navsim_v2_main}
    \scriptsize
    \resizebox{\linewidth}{!}{%
    \begin{tabular}{lcccccccccc}
        \toprule
        Method & NC & DAC & DDC & TLC & EP & TTC & LK & HC & EC & EPDMS \\
        \midrule
        \textit{ReCogDrive~\cite{xiong2026recogdrive}}
            & 98.3 & 95.2 & 99.5 & 99.8 & 87.1 & 97.5 & 96.6 & 98.3 & 86.5 & 83.6 \\
        \textit{DiffusionDriveV2*~\cite{diffusiondrivev2}}
            & 97.7 & 96.6 & 99.2 & 99.8 & 88.9 & 97.2 & 96.0 & 97.8 & 91.0 & 85.5 \\
        \textit{DriveVLA-W0*~\cite{drivevla}}
            & 98.5 & 99.1 & 98.0 & 99.7 & 86.4 & 98.1 & 93.2 & 97.9 & 58.9 & 86.1 \\
        \textit{DriveSuprim (ViT-L)~\cite{yao2026drivesuprim}}
            & 98.4 & 98.6 & 99.6 & 99.8 & 90.5 & 97.8 & 97.0 & 98.3 & 78.6 & 87.1 \\
        \textit{ExploreVLA*~\cite{explorevla}}
            & 98.8 & 96.2 & 99.6 & 99.8 & 87.1 & 98.2 & 97.8 & 98.3 & 86.8 & 88.8 \\
        \textit{DriveFine†~\cite{dang2026drivefine}}
            & 98.7 & 97.3 & 99.5 & 99.8 & 88.7 & 97.8 & 97.7 & 98.4 & 83.8 & 89.7 \\
        \midrule
        \textbf{RefineDrive†}
            & 98.7 & 98.2 & 98.1 & 99.7 & 90.5 & 98.2 & 94.4 & 97.8 & 82.7 & 89.4 \\
        \bottomrule
    \end{tabular}%
    }
\end{table}

\subsection{Ablation Studies}
\label{sec:ablation}

We conduct controlled ablations on NAVSIM v1 to study:
(1) learning from self-generated failures;
(2) minimal correction versus Human GT;
(3) the benefit of structured failure-diagnosis supervision; and
(4) Safety-Layered Reward versus direct PDMS optimization.
All variants share the same base SFT model and evaluation protocol, and
Human-GT and minimal-correction variants use the same failure cases.


\begin{table}[t]
    \centering
    \setlength{\belowcaptionskip}{6pt}
    \caption{
        Ablation of failure-aware Correction SFT.
        Human GT and Minimal use the same failure cases and identical
        structured diagnosis supervision.
        Minimal w/o Diagnosis retains the same corrective trajectory
        targets as Minimal but removes diagnosis supervision.
    }
    \label{tab:ablation_correction}
    \scriptsize
    \begin{tabular}{lcccccc}
        \toprule
        Correction Strategy & NC & DAC & EP & TTC & C & PDMS \\
        \midrule
        Base SFT              & 98.6 & 95.8 & 81.1 & 95.4 & \textbf{100} & 87.7 \\
        Human GT              & 98.6 & 96.1 & 81.6 & 95.7 & \textbf{100} & 88.2 \\
        Minimal w/o Diagnosis & 98.5 & 96.2 & 81.7 & 95.7 & \textbf{100} & 88.3 \\
        Minimal               & \textbf{98.7} & \textbf{96.4} &
                                \textbf{82.0} & \textbf{95.9} & \textbf{100} &
                                \textbf{88.6} \\
        \bottomrule
    \end{tabular}
\end{table}

\paragraph{Failure-Aware Correction.}
Using self-generated failures with Human GT supervision improves NAVTEST PDMS from 87.7 to 88.2.
With the failure cases and structured diagnosis supervision held fixed,
replacing Human GT with retrieved minimal corrections further increases PDMS to 88.6, with gains in NC, DAC, EP, and TTC.
Across the 32,650 paired Correction-SFT records, however, the retrieved
targets have a slightly lower mean PDMS than the corresponding Human GT trajectories (95.4 vs.\ 95.8).
Thus, the downstream improvement cannot be explained by higher average target PDMS. Furthermore, removing diagnosis supervision while keeping the corrective
trajectory targets unchanged reduces PDMS from 88.6 to 88.3.
This suggests that structured failure-diagnosis supervision benefits
the resulting planning policy beyond corrective trajectory supervision alone.

\begin{table}[t]
    \centering
    \setlength{\belowcaptionskip}{6pt}
    \caption{
        Ablation of the RL objective from the same minimal-correction SFT checkpoint.
    }
    \label{tab:ablation_reward}
    \scriptsize
    \begin{tabular}{lcccccc}
        \toprule
        RL Objective & NC & DAC & EP & TTC & C & PDMS \\
        \midrule
        No RL                 & \textbf{98.7} & 96.4 & 82.0 &
                                \textbf{95.9} & \textbf{100} & 88.6 \\
        PDMS Reward           & 98.2 & 97.8 & \textbf{87.2} &
                                94.9 & \textbf{100} & 91.1 \\
        Safety-Layered Reward & \textbf{98.7} & \textbf{98.2} &
                                \textbf{87.2} & \textbf{95.9} & \textbf{100} &
                                \textbf{91.7} \\
        \bottomrule
    \end{tabular}
\end{table}

\paragraph{Safety-Layered Reinforcement Learning.}
Table~\ref{tab:ablation_reward} isolates the RL objective from the same
minimal-correction SFT checkpoint.
Direct PDMS optimization increases EP from 82.0 to 87.2,
but lowers NC from 98.7 to 98.2 and TTC from 95.9 to 94.9.
Our Safety-Layered Reward reaches the same reported mean
EP (87.2), while retaining the pre-RL NC and TTC scores
and increasing DAC to 98.2.
Relative to PDMS Reward, it improves NC, DAC, and TTC by
0.5, 0.4, and 1.0 percentage points, respectively,
and PDMS from 91.1 to 91.7.
This comparison supports improved safety without a reduction
in reported mean progress.


\begin{table}[!h]
    \centering
    \setlength{\belowcaptionskip}{6pt}
    \caption{
        Effect of Correction SFT under the same Safety-Layered GRPO objective. Full Correction SFT achieves the highest NC, DAC, and TTC among the compared variants, with lower EP.
    }
    \label{tab:ablation_full}
    \scriptsize
    \begin{tabular}{lcccccc}
        \toprule
        Training Variant & NC & DAC & EP & TTC & C & PDMS \\
        \midrule
        w/o Correction SFT
            & 97.8 & 97.7 & 88.8 & 94.1 & \textbf{100} & 91.2 \\
        Minimal-Correction SFT w/o Diagnosis
            & 98.2 & 97.8 & 87.5 & 95.3 & 99.9 & 91.3 \\
        Human-GT Correction SFT
            & 98.1 & 97.7 & \textbf{89} & 94.7 & 99.9 & 91.6 \\
        Full Correction-SFT
            & \textbf{98.7} & \textbf{98.2} & 87.2 &
              \textbf{95.9} & \textbf{100} & \textbf{91.7} \\
        \bottomrule
    \end{tabular}
\end{table}

\paragraph{Effect on the Final Policy.}
Table~\ref{tab:ablation_full} evaluates Correction SFT variants followed by
the same Safety-Layered GRPO.
Full Correction SFT attains the highest NC, DAC, and TTC
among all compared variants.
Relative to omitting Correction SFT, the gains are
0.9, 0.5, and 1.8 percentage points, respectively,
with EP decreasing from 88.8 to 87.2.
Relative to Human-GT Correction SFT, it increases the three
safety metrics by 0.6, 0.5, and 1.2 points, while EP decreases
from 89.0 to 87.2.
Thus, their similar PDMS values (91.7 vs. 91.6) accompany
different safety--progress profiles and do not imply
similar safety performance.
Removing diagnosis while retaining the same corrective
targets lowers NC, DAC, and TTC by 0.5, 0.4, and 0.6 points,
with EP increasing from 87.2 to 87.5.
Together, these comparisons support a safety-oriented
contribution from diagnosis-and-correction supervision
after RL, with an explicit progress trade-off.

\subsection{Qualitative Analysis}

\begin{figure*}[!t]
    \centering

    \setlength{\tabcolsep}{3pt}
    \renewcommand{\arraystretch}{1.0}

    \begin{tabular}{@{}ccccc@{}}
        &
        \textbf{Scene} &
        \shortstack{\textbf{Initial}\\\textbf{Prediction}} &
        \textbf{RefineDrive} &
        \textbf{Human GT}
        \\[8pt]


        \adjustbox{valign=m}{\textbf{(a)}} &
        \adjustbox{valign=m}{
            \includegraphics[
                height=0.17\textwidth,
                keepaspectratio
            ]{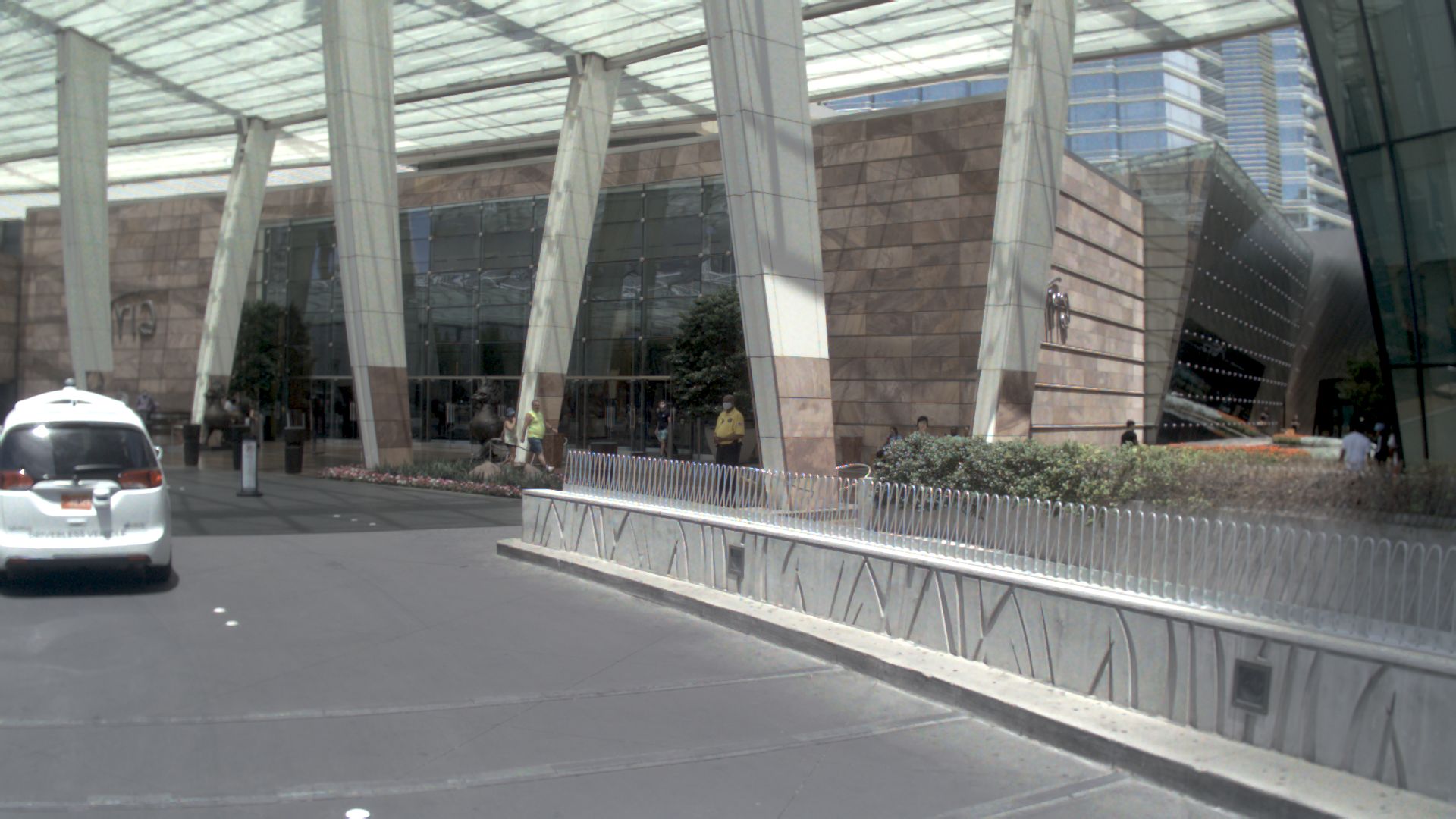}
        } &
        \adjustbox{valign=m}{
            \includegraphics[
                height=0.17\textwidth,
                keepaspectratio
            ]{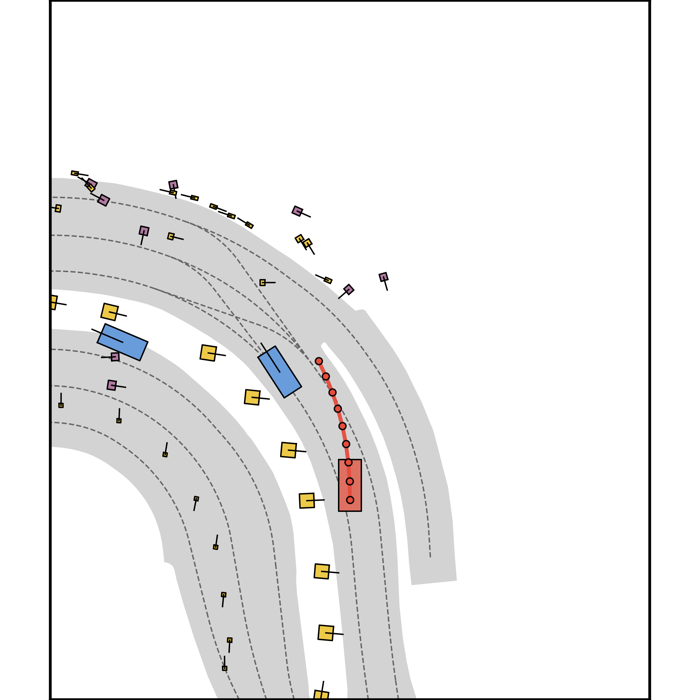}
        } &
        \adjustbox{valign=m}{
            \includegraphics[
                height=0.17\textwidth,
                keepaspectratio
            ]{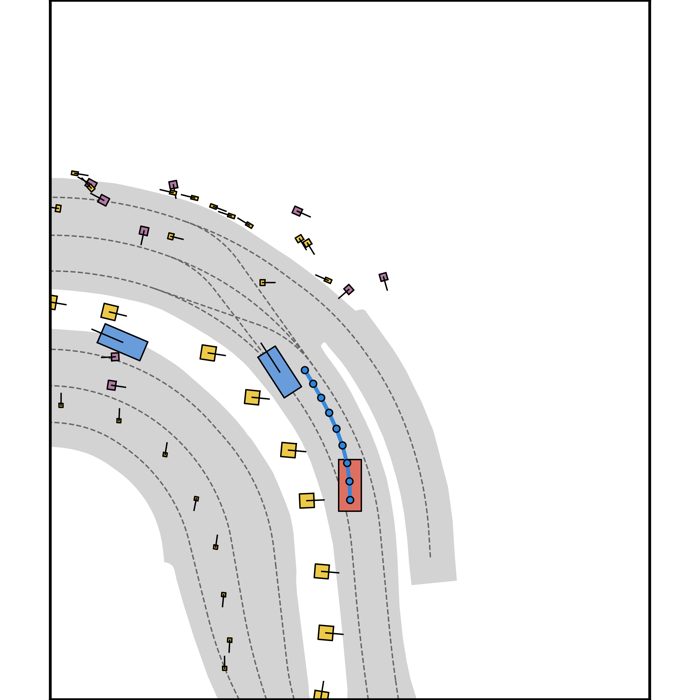}
        } &
        \adjustbox{valign=m}{
            \includegraphics[
                height=0.17\textwidth,
                keepaspectratio
            ]{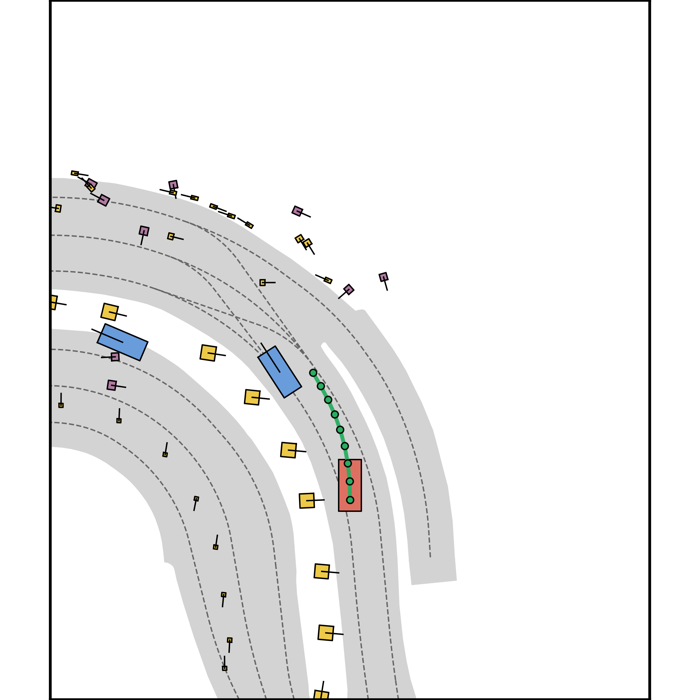}
        }
        \\[35pt]

        \adjustbox{valign=m}{\textbf{(b)}} &
        \adjustbox{valign=m}{
            \includegraphics[
                height=0.17\textwidth,
                keepaspectratio
            ]{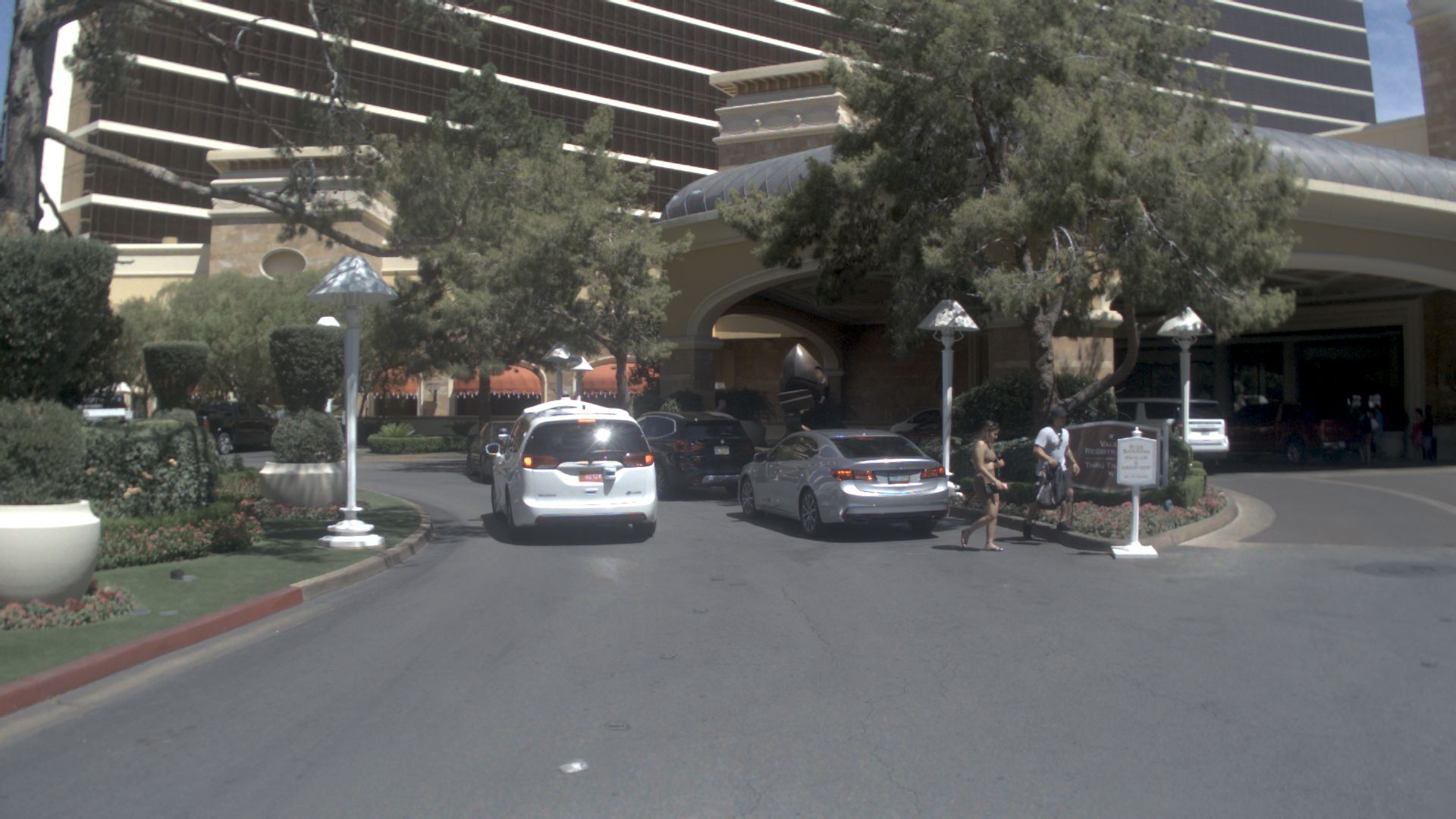}
        } &
        \adjustbox{valign=m}{
            \includegraphics[
                height=0.17\textwidth,
                keepaspectratio
            ]{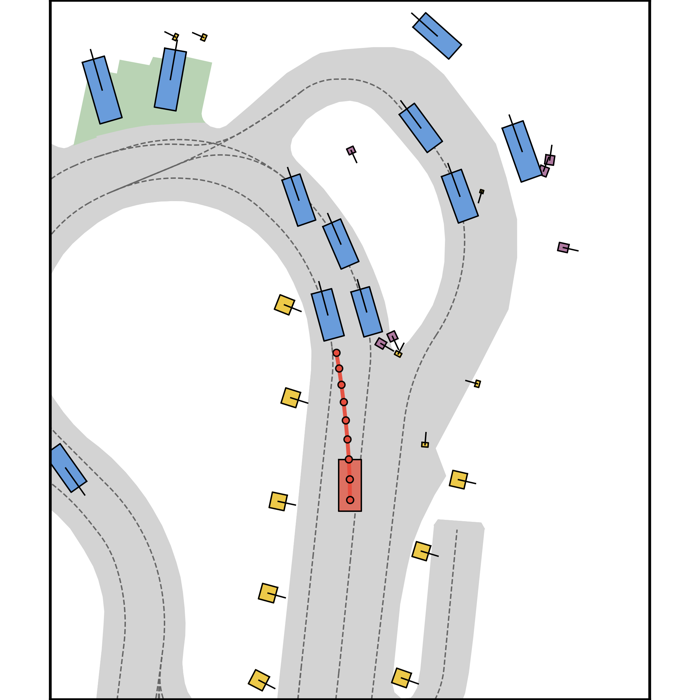}
        } &
        \adjustbox{valign=m}{
            \includegraphics[
                height=0.17\textwidth,
                keepaspectratio
            ]{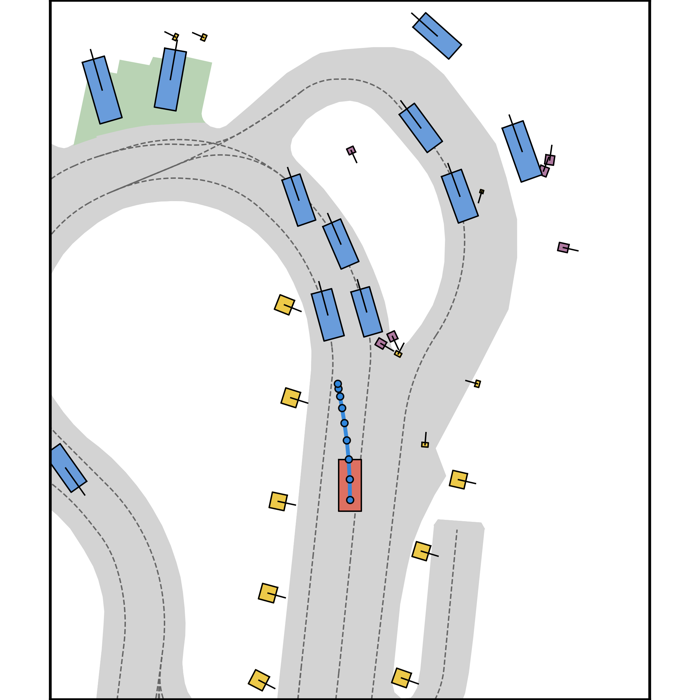}
        } &
        \adjustbox{valign=m}{
            \includegraphics[
                height=0.17\textwidth,
                keepaspectratio
            ]{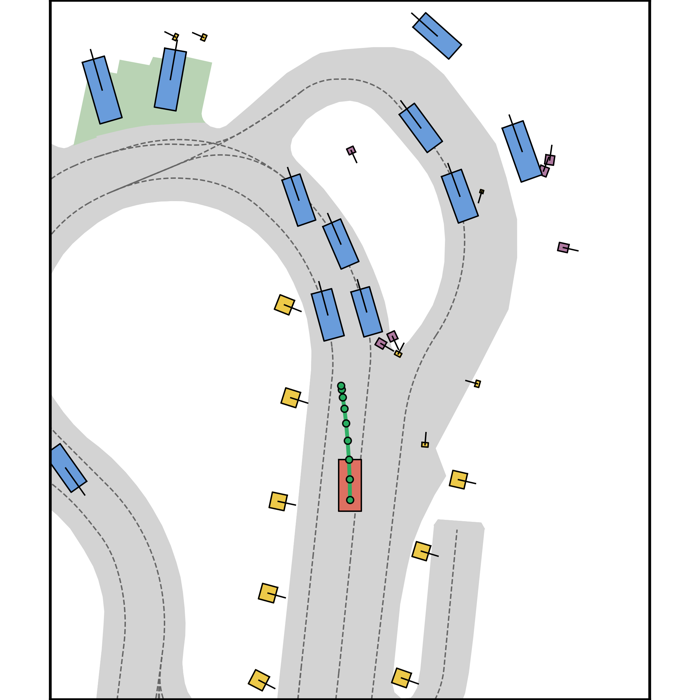}
        }

    \end{tabular}

    \caption{
        Qualitative comparison of direct trajectory predictions on NAVTEST:
        (a) off-road cases and (b) collision cases.
        Red, blue, and green denote the base-policy prediction,
        RefineDrive prediction, and human trajectory, respectively.
        Both policies predict independently from the same driving context;
        RefineDrive is not conditioned on the base-policy prediction.
    }
    \label{fig:qualitative}
\end{figure*}

Fig.~\ref{fig:qualitative} compares the base SFT policy and RefineDrive
on the same NAVTEST scenes.
Both policies directly predict trajectories from the driving context.
In these examples, RefineDrive avoids the off-road and collision
behaviors exhibited by the base policy.

%% file: sections/conclusion.tex
\section{Conclusion}
\label{sec:conclusion}

We present RefineDrive, a failure-guided post-training framework integrating Reliable Diagnosis, Minimum-Correction Target Retrieval, and Safety-Layered GRPO, with no inference-time diagnosis or repair. Controlled ablations show that diagnosis-and-correction supervision improves NC, DAC, and TTC after RL at some cost to progress, while Safety-Layered Reward improves all three at the same mean EP as direct PDMS optimization. RefineDrive achieves 91.7 PDMS on NAVSIM v1 and, with the same checkpoint and no further training, 89.4 EPDMS on the original NAVTEST scenes under NAVSIM v2 extended metrics.